\documentclass[runningheads]{llncs}

\usepackage[T1]{fontenc}
\usepackage{graphicx}
\usepackage{amsmath,amssymb,amsfonts}
\usepackage{booktabs}
\usepackage{multirow}
\usepackage{xcolor}
\usepackage{hyperref}
\usepackage{cleveref}
\usepackage{subcaption}
\usepackage{adjustbox}

\begin{document}

\title{Toward Aristotelian Medical Representations: Backpropagation-Free Layer-wise Analysis for Interpretable Generalized Metric Learning on MedMNIST}

\titlerunning{Aristotelian Representations for Interpretable Metric Learning}

\author{Michael Karnes\inst{1} \and
Alper Yilmaz\inst{1}}

\authorrunning{M. Karnes and A. Yilmaz}

\institute{The Ohio State University, Columbus, OH 43210, USA\\
\email{\{karnes.30, yilmaz.15\}@osu.edu}}

\maketitle              

\begin{abstract}
While deep learning has achieved remarkable success in medical imaging, the "black-box" nature of backpropagation-based models remains a significant barrier to clinical adoption. To bridge this gap, we propose Aristotelian Rapid Object Modeling (A-ROM), a framework built upon the Platonic Representation Hypothesis (PRH). This hypothesis posits that models trained on vast, diverse datasets converge toward a universal and objective representation of reality. By leveraging the generalizable metric space of pretrained Vision Transformers (ViTs), A-ROM enables the rapid modeling of novel medical concepts without the computational burden or opacity of further gradient-based fine-tuning. We replace traditional, opaque decision layers with a human-readable concept dictionary and a $k$-Nearest Neighbors ($k$NN) classifier to ensure the model's logic remains interpretable. Experiments on the MedMNIST v2 suite demonstrate that A-ROM delivers performance competitive with standard benchmarks while providing a simple and scalable, "few-shot" solution that meets the rigorous transparency demands of modern clinical environments.

\end{abstract}

\section{Introduction}
Deep neural networks (DNNs) have revolutionized data processing, pattern recognition, and, more recently, generative modeling. However, as these models are integrated into sensitive domains, critical questions have emerged regarding their computational overhead, generalizability, and inherent opacity. While the success of large language models has demonstrated significant generalizability in text, the Platonic Representation Hypothesis (PRH) \cite{huh2024platonic} suggests that large-scale networks are converging toward a shared, universal statistical model of reality across both textual and visual modalities.

Historically, researchers have exploited shared visual patterns through adaptation techniques such as fine-tuning, transfer learning, and meta-learning. Yet, these frameworks often inherit the limitations of their specific training data and the stochastic ambiguity introduced by backpropagation, which frequently results in high architectural complexity. To overcome these limitations, we introduce Aristotelian Rapid Object Modeling (A-ROM). By leveraging the universal metric space described by the PRH, A-ROM enables rapid visual learning via a transparent, interpretable classification process. This approach reflects the Aristotelian view that human knowledge emerges from the organization of experience via innate cognitive primitives. A-ROM mimics this process, utilizing the PRH's universal latent features as a template to synthesize complex medical imagery into structured conceptual representations.

We demonstrate the efficacy of this framework using MedMNIST v2 \cite{yang2023medmnist}, a suite selected for its diversity and its relevance to high-stakes medical decision-making. The following sections detail recent work addressing clinical AI challenges and the methodology of the A-ROM framework. We then provide a layer-wise analysis of the pretrained DINOv2 ViT-Large architecture \cite{oquab2023dinov2}, incorporating a direct performance comparison against established benchmarks within this exploration. Finally, we evaluate the framework’s potential through a few-shot learning experiment and discuss the broader applications of A-ROM in environments requiring rapid, online adaptation and high interpretability.

\section{Related Work}
\label{sec:related_work}
\subsection{The Platonic Foundation: Universal Metric Convergence}
The A-ROM framework is predicated on the PRH, which posits that the latent manifolds of deep networks converge toward a shared, objective geometry, regardless of architecture or training task \cite{huh2024platonic}. This convergence suggests that an optimal representation is a statistical destination rather than a task-specific accident \cite{ziyin2025proof}. The existence of this shared geometry is empirically supported by the "stitchability" of disparate architectures; latent layers from distinct models, such as ViTs and CLIP, can be bridged via simple affine transformations with negligible performance loss. This suggests that the relative geometry between data points remains remarkably consistent across the frontier of AI research \cite{lenc2015understanding,huh2024platonic,moschella2023relative}. As models scale, they converge toward a shared internal language that renders their feature spaces functionally interchangeable \cite{bansal2021revisiting}.

This artificial convergence mirrors biological evolution, implying a "canonical" organization of information shared by both artificial and natural intelligence. Historically, unsupervised algorithms tasked with sparse coding spontaneously developed receptive fields resembling the Gabor filters of the primary visual cortex \cite{olshausen1996emergence}. Modern scaling laws confirm this trajectory, as foundation models exhibit mid-to-late layer structures that align closely with human sensory and prefrontal cortex activity \cite{simeoni2025disentangling,lopezcardona2025brainlanguage}.

The degree of a model's alignment with this "Platonic ideal" serves as a direct predictor of few-shot generalization capabilities \cite{lu-etal-2025-representation}. This grounding is vital under extreme data scarcity, as it allows A-ROM to rely on the pre-existing structural integrity of the converged manifold rather than task-specific retraining \cite{ji2025optimal,karnes2024key}. Consequently, A-ROM treats the frozen backbone not as a black-box feature extractor, but as a structured, universal manifold. By anchoring to these stable distributions, the framework captures fundamental structural regularities, enabling robust classification even when provided with minimal clinical examples.

\subsection{Interpretability and Clinical AI}
\subsubsection{Explainable AI}
High-stakes sectors like finance, law, and cybersecurity increasingly mandate Explainable AI (XAI) to mitigate the systemic risks of "black-box" decision-making \cite{mohsin2025explaining,EU_AI_Act_2024}. Beyond legal compliance, transparency is a functional necessity, often achieved through post-hoc interpretability. Methods such as SHAP and LIME provide feature-importance scores to justify individual predictions, while spatial tools like Grad-CAM utilize gradients to generate retroactive heatmaps as proxies for a model's focus \cite{selvaraju2017gradcam}.

Building on post-hoc foundations, the field of autonomous driving is advancing toward ante-hoc, or intrinsic, interpretability \cite{ugboko2025avxai}. This forward-looking paradigm moves beyond retroactive auditing by embedding transparency directly into the model's architecture. By integrating saliency maps and counterfactual reasoning into the design phase, these systems replace external summaries with structural transparency.

\subsubsection{Clinical XAI: Utility, Gaps, and Regulatory Mandates} 
A similar evolution is unfolding in clinical diagnostics, where the shift from static prediction to active reasoning has driven the use of post-hoc tools like SHAP and Grad-CAM++ to identify physiological and morphological risk markers \cite{guattery2025opioid,gao2025knee,jain2025interpretable}. However, these methodologies often face a "fidelity gap," as saliency-based explainers struggle with the complex textures of pathology and lack the causal depth required for clinical trust \cite{singh2025beyond,Carriero2025}.

This technical challenge is undercored by a critical regulatory shift: the 2025 FDA draft guidance \cite{fda2025guidance} now mandates context-specific validation and immutable audit trails for high-risk AI. Consequently, the field is moving away from retroactive proxies toward traceable, "interpretable-by-design" architectures. In this high-stakes environment, intrinsic transparency has become a functional requirement for both regulatory accountability and patient safety \cite{hulsen2023xai,Ennab2024}.

\subsubsection{Concept-Centric Interpretability}
A critical distinction exists between explainability, which relies on post-hoc surrogates, and interpretability, where the architecture is understandable by design \cite{rudin2019stop,cao2024enhancing}. In a seminal critique, \cite{rudin2019stop} argues that high-stakes decisions should prioritize such inherent interpretability over retroactive "guessing." This has catalyzed "interpretable-by-design" frameworks that utilize a "Concept Dictionary" to map latent activations to symbolic features, such as vessel tortuosity or nuclei density \cite{huy2025interactive,corbetta2025inhoc}. While these models provide a descriptive "visual vocabulary," they often lack the situational context inherent in case-based evidence.

A-ROM addresses this limitation by replacing opaque decision layers with distance-based logic, shifting justification from abstract probabilities to localized, neighbor-based evidence \cite{papernot2018deep}. By anchoring decisions to the stable distributions described by the PRH, the framework prioritizes universal data regularities over dataset-specific noise. This facilitates a "human-in-the-loop" workflow where clinicians audit diagnostic paths via retrieved, peer-validated exemplars. Ultimately, A-ROM transforms opaque inferences into a transparent evidentiary chain suitable for high-stakes clinical integration.

\subsection{The MedMNIST Benchmark Ecosystem}
\label{subsec:medmnist_ecosystem}
The MedMNIST v2 suite \cite{yang2023medmnist}, an expansion of the foundational v1 release \cite{yang2021medmnist}, has emerged as a standardized "decathlon" for medical image analysis. It comprises 12 2D and 6 3D datasets spanning the primary modalities of modern clinical practice, including X-ray, Optical Coherence Tomography (OCT), Ultrasound, Computed Tomography (CT), and Electron Microscopy. By providing pre-processed, high-quality data across such a diverse task spectrum, the suite enables a rigorous evaluation of how effectively general-purpose features translate to specialized medical domains. MedMNIST offers a rigorous testing ground to prove that A-ROM’s universal features effectively translate to specialized medical imaging.


\begin{figure}[ht]
    \centering
    \includegraphics[width=1.0\textwidth]{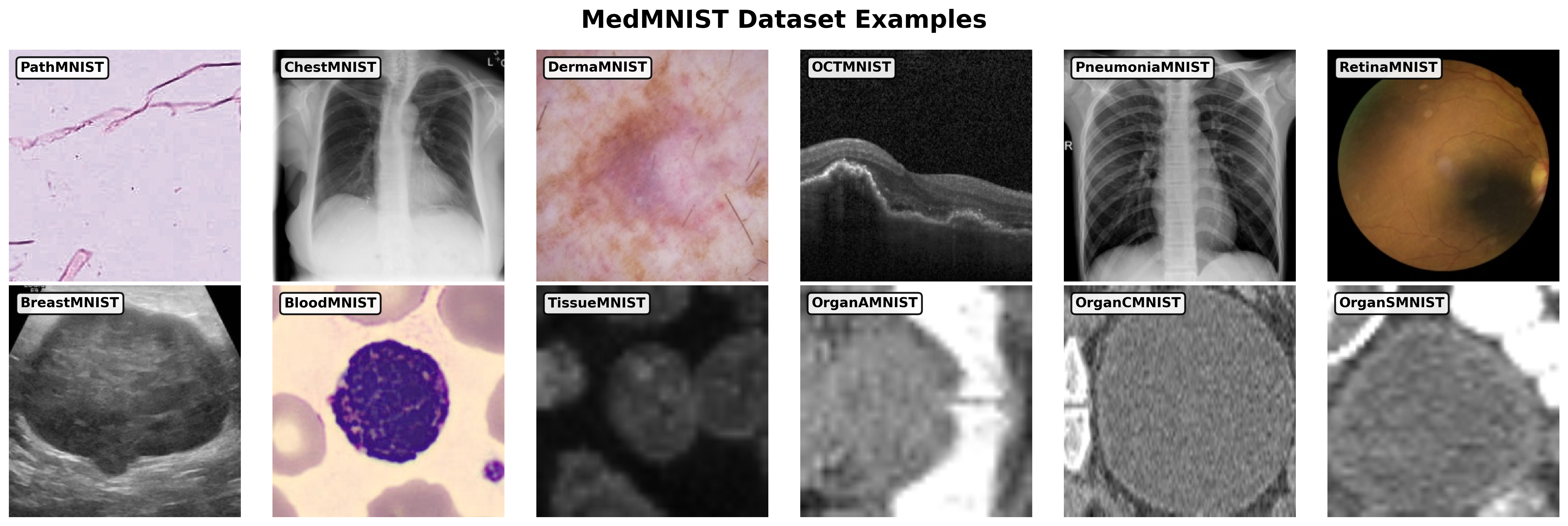}
    \caption{Visual overview of the MedMNIST v2 benchmark, featuring sample images from each of the 12 constituent 2D datasets \cite{yang2023medmnist}.}
    \label{fig:medmnist}
\end{figure}

\subsubsection{MedMNIST as a Prototyping Benchmark for Architectures}
Recent research using MedMNIST has increasingly relied on complex, task-specific refinements to stabilize latent geometry. Approaches range from Center Loss clustering \cite{zeng2022learning} and Supervised Contrastive Learning \cite{mildenberger2025tale} to adversarial distance minimization \cite{medicat2024} and Earth Mover’s Distance for few-shot tasks \cite{wu2025mrwvit}. While these innovations yield performance gains, they entail sensitive hyperparameter tuning and significant computational overhead. This trend toward intricate pipelines underscores the need for A-ROM’s minimalist, low-training architecture.

Within this landscape, MedMNIST has become a critical benchmark for investigating how architectural complexity translates into clinical utility. For instance, \cite{barekatain2025evaluating} established a baseline for explainability by evaluating ViTs against CNNs using localized saliency proxies. Similarly, \cite{doerrich2025rethinking} recently re-evaluated prototype-based logic, comparing end-to-end training and linear probing against $k$NN evaluations. However, while \cite{doerrich2025rethinking} identifies which backbones best preserve prototypes across multiple image resolutions, their analysis treats prototyping primarily as an output-layer phenomenon and focuses its discussion on global MedMNIST averages rather than exploring the unique performance trends of individual datasets.

Our work extends these findings by investigating the efficacy of the PRH and exploring the evolution of latent features through the internal layers of the DINOv2 ViT-Large architecture \cite{oquab2023dinov2} for $k$NN classification on MedMNIST, incorporating dimensionality reduction to maintain the topological structure required for both scalability and practical efficiency.

\section{Methodology}
\label{sec:method}
The A-ROM framework utilizes a staged pipeline that decouples feature extraction from class-specific modeling inspired by \cite{karnes2025rapid}. Stage 1 performs an unsupervised distillation of Platonic ideals, extracting innate, universal latent features to form an 'encoding language' that preserves the topological relationships of the embedding space. Stage 2 executes Aristotelian concept formation, using supervised alignment to map these structural regularities into a concept dictionary. This architecture enables generalizability by grounding abstract, high-dimensional universal forms into structured, categorical knowledge.

\subsection{Stage 1: Unsupervised Encoding of Platonic Ideals}
To distill these Platonic ideals, the framework derives a compressed encoding language from unlabeled images by extracting latent features from transformer block $\ell$ of a frozen DINOv2-Large backbone \cite{oquab2023dinov2}. For an input $x$, the global latent vector $z \in \mathbb{R}^{1024}$ is obtained by averaging $N=256$ patch tokens:\begin{equation}z = \frac{1}{N} \sum_{i=1}^{N} p_{i, \ell}\end{equation}where $p_{i, \ell}$ represents the $i$-th patch token. To form the Alphabet, Principal Component Analysis (PCA) projects the centered latent vector into a reduced space via:\begin{equation}a = W_{\text{PCA}}^T (z - \mu)\end{equation}This space is quantized into a Vocabulary via $K$-means clustering into a set of $V$ centroids $\mathcal{V} = \{v_1, \dots, v_V\}$. We define the Word Vector $d \in \mathbb{R}^V$ based on the Euclidean distances to each of these centroids, where $d_j = \|a - v_j\|_2$. The final Full Encoding vector $s$ is constructed by concatenating the Alphabet vector, produced by the PCA transform, with this Word vector:\begin{equation}s = [a \oplus d]\end{equation}

\subsection{Stage 2: Supervised Synthesis of Aristotelian Concepts}
To synthesize these distilled features into Aristotelian concepts, a labeled dataset is used to define class-specific regions and calculate a final Linear Discriminant Analysis (LDA) transform. Labeled training samples for each class $c$ generate a collection $\mathcal{S}_c = \{s_{1,c}, \dots, s_{n,c}\}$. To support discriminant alignment, the empirical covariance $\Sigma_c$, within-class scatter $S_W$, and between-class scatter $S_B$ are calculated:

\begin{equation}\Sigma_c = \frac{1}{n_c-1} \sum_{i=1}^{n_c} (s_{i,c} - \bar{s}_c)(s_{i,c} - \bar{s}_c)^T\end{equation}

\begin{equation}S_W = \sum_{c} \Sigma_c, \quad S_B = \sum_{c} n_c (\bar{s}_c - \bar{s})(\bar{s}_c - \bar{s})^T\end{equation}

where $\bar{s}_c$ is the class mean and $\bar{s}$ is the global mean. The LDA projection matrix $W_{\text{LDA}}$ is obtained by solving the generalized eigenvalue problem $S_B w = \lambda S_W w$. This optimization ensures that the structural regularities of the encoding language are aligned along axes of maximum class distinction, effectively maximizing the ratio of between-class variance to within-class variance.

\subsection{Stage 3: Inference via Per-Class Mahalanobis Distance}
During inference, a test image $x_{\text{test}}$ is encoded into $s_{\text{test}}$ and projected into the discriminant space: $\tilde{s}_{\text{test}} = W_{\text{LDA}}^T s_{\text{test}}$. The sample is evaluated against every exemplar $\tilde{e}_c$ in the dictionary using the Mahalanobis metric, normalized by the projected covariance $\tilde{\Sigma}_c$:
\begin{equation}
D_M(\tilde{s}_{\text{test}}, \tilde{e}_c) = \sqrt{(\tilde{s}_{\text{test}} - \tilde{e}_c)^T \tilde{\Sigma}_c^{-1} (\tilde{s}_{\text{test}} - \tilde{e}_c)}
\end{equation}
The predicted label $\hat{y}$ is assigned via a majority vote among the $k$ nearest neighbors:
\begin{equation}
\mathcal{N}_k(\tilde{s}_{\text{test}}) = \arg\min_{e \in \mathcal{D}_{\text{train}}}^{(k)} D_M(\tilde{s}_{\text{test}}, \tilde{e}_c)
\end{equation}
This is what provides the identifiable evidentiary chain for the final classification.

\section{Experimental Design}
This study evaluates the A-ROM framework utilizing a DINOv2-ViT-L/14 backbone \cite{oquab2023dinov2}. The experimental pipeline follows a structured progression from hyperparameter optimization to large-scale benchmarking and few-shot analysis. 

\subsection{Hyperparameter Optimization and Benchmarking}
The framework was first subjected to a coarse-to-fine parameter sweep across 11 of the 12 2D datasets of the MedMNIST v2 suite (224 $\times$ 224 resolution). Due to its multi-label nature, ChestMNIST was omitted to ensure architectural consistency with our distance-based, single-label classification pipeline. This sweep evaluated the interaction between the network layer $\ell$ and two key hyperparameters: the Alphabet size ($A \in \{64, 256, 512\}$ components) and the Vocabulary size ($V \in \{64, 256, 512\}$ clusters).

To maintain computational tractability across the high volume of trials, the initial sweep utilized 1,000 training images for language construction, 64 training images per class for the concept dictionary, and 200 validation images for evaluation ($k=3$). A second, fine-grained sweep followed, fixing the optimal network layer to search the localized neighborhood of the highest-performing Alphabet and Vocabulary sizes.

Using these optimized parameters, A-ROM was evaluated against the full test sets of all the 11 considered MedMNIST datasets using $k=15$ nearest neighbors. During this benchmarking stage, training labels were capped at 5,000 per class for both the encoding language and dictionary construction.

\subsection{Label Efficiency and Few-Shot Robustness}
The final investigation focused on the impact of label availability on diagnostic performance. The encoding language was fixed using the benchmarking configuration (up to 5,000 samples per class). The sample size utilized for the supervised concept dictionary was then varied from 8 to 512 per class, randomly drawn across five independent repeats. Each trial was evaluated against the full test sets for the 11 considered MedMNIST datasets using $k=15$.

\section{Results}
\label{sec:selection}

\subsection{Hyperparameter Sensitivity and Layer-wise Performance} 
\begin{figure}[ht]
    \centering
    \includegraphics[width=1.0\textwidth]{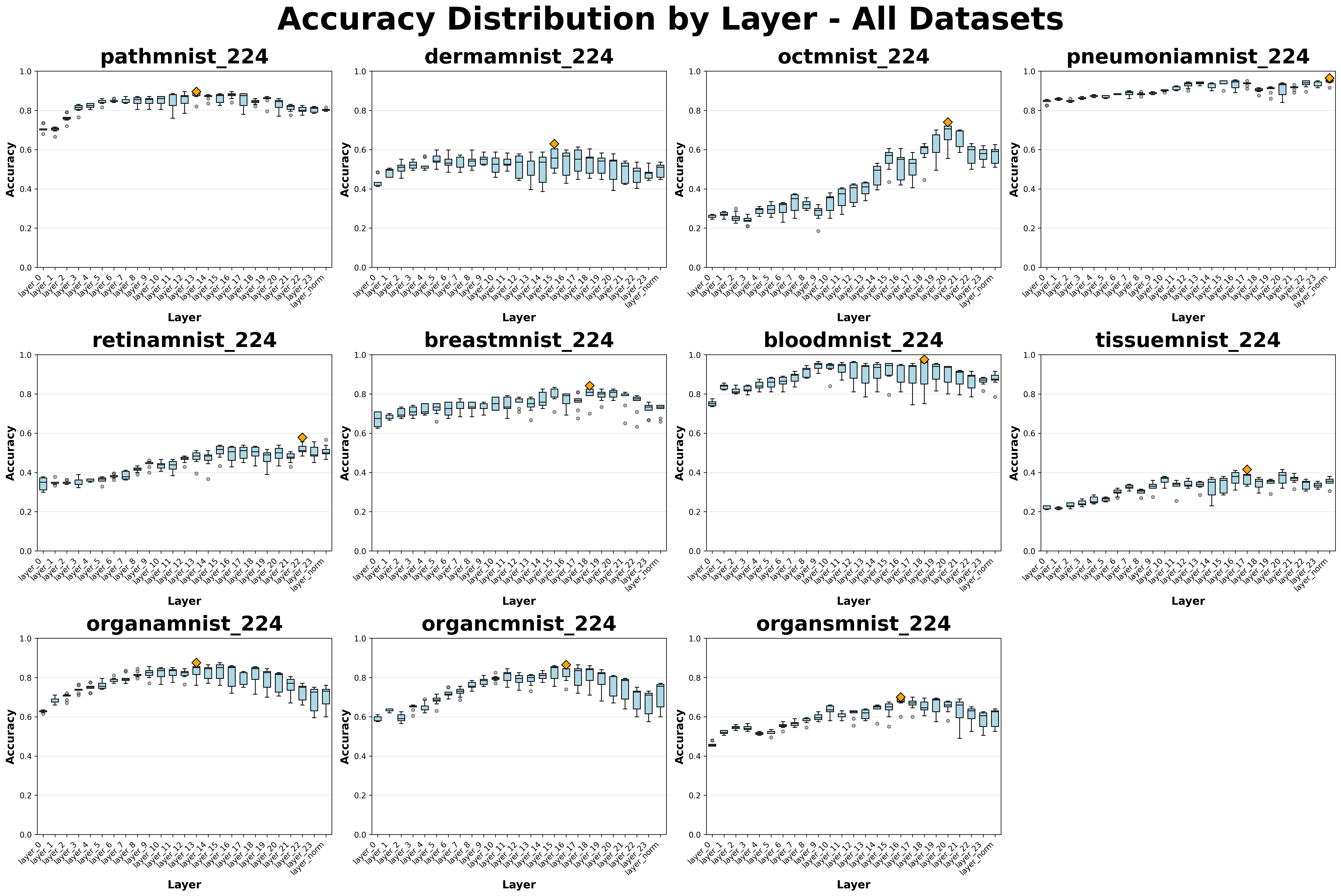}
    \caption{Layer-wise classification performance across the 25 transformer blocks of the DINOv2-L/14 backbone. Each box plot aggregates the accuracy variance for a specific depth across 11 MedMNIST v2 datasets, reflecting the interplay between different Alphabet and Vocabulary parameter combinations. Peak accuracies for each layer are indicated by orange markers.}
    \label{fig:layer_sweep}
\end{figure}

Figure \ref{fig:layer_sweep} illustrates the classification accuracies across 25 layers of the DINOv2 backbone for the 11 considered MedMNIST v2 datasets. This analysis incorporates nine parameter combinations of Alphabet and Vocabulary sizes for each layer, revealing several distinct trends in model response.

A prominent "mound-like" trend is observed across most datasets, where the highest accuracies are concentrated within the middle layers of the network. This suggests that intermediate representations offer the optimal balance between low-level structural features and high-level semantic abstractions. Conversely, deeper layers exhibit a wider range of accuracies, indicating a heightened sensitivity to Alphabet and Vocabulary sizes as the feature space becomes more specialized.

The most distinct behavior is observed in the OCTMNIST dataset, which demonstrates a sharp performance peak in the deeper layers followed by a precipitous decline in the final blocks. This suggests that for certain specialized medical modalities, the choice of layer depth is more critical than for generalized tasks. Overall, datasets that achieved high peak accuracies maintained relatively high performance across all layers, while inherently difficult datasets exhibited consistently lower performance regardless of depth.

The optimal layer depth, Alphabet size, and Vocabulary size identified during the refinement sweep are summarized in Table \ref{tab:params}. Most datasets reached peak performance within the mid-to-late blocks (layers 13 to 18), while only one dataset achieved its optimal results using the final layer. Regarding the Alphabet size, the 11 datasets bifurcate into two distinct groups: those requiring approximately 512 components and those optimized at roughly 256. Notably, the majority of datasets reached peak accuracy using fewer than 100 clusters. This represents an significant reduction in dimensionality from the original 1024-dimensional latent vector $z$, demonstrating the framework's ability to maintain high diagnostic performance while significantly compressing the underlying feature space.

\begin{table}[htbp]
\centering
\caption{Optimal Hyperparameters across MedMNIST Datasets}
\label{tab:medmnist_results}
\begin{adjustbox}{width=\columnwidth}
\begin{tabular}{@{}lcccccccccccc@{}}
\toprule
\textbf{Metric} & \textbf{Path}  & \textbf{Derma} & \textbf{OCT} & \textbf{Pneumo} & \textbf{Retina} & \textbf{Breast} & \textbf{Blood} & \textbf{Tissue} & \textbf{OrganA} & \textbf{OrganC} & \textbf{OrganS} \\ \midrule
Layer & $\ell$-13  & $\ell$-15 & $\ell$-20 & $\ell$-final & $\ell$-22 & $\ell$-18 & $\ell$-18 & $\ell$-17 & $\ell$-13 & $\ell$-16 & $\ell$-16 \\
Components & 224  & 512 & 512 & 224 & 488 & 248 & 496 & 512 & 248 & 248 & 272 \\
Clusters & 56 & 88 & 48 & 480 & 56 & 32 & 488 & 288 & 288 & 72 & 96 \\ \bottomrule
\end{tabular}
\end{adjustbox}
\label{tab:params}
\end{table}

\begin{figure}[ht]
    \centering
        \includegraphics[width=1.0\textwidth]{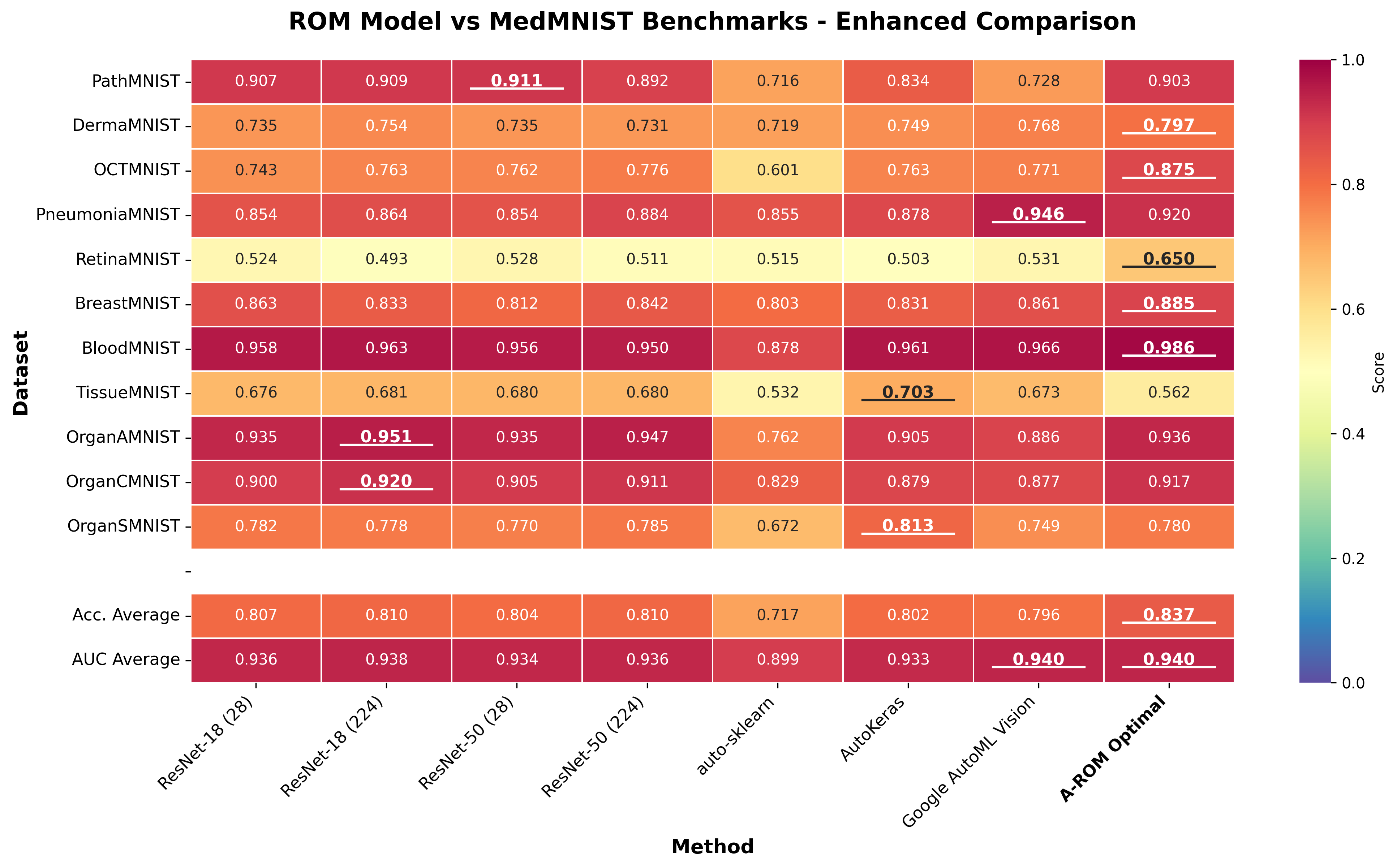}
    \caption{Performance heatmap of A-ROM versus established MedMNIST v2 benchmarks \cite{yang2023medmnist} for the 11 considered datasets. The comparison evaluates the optimized configuration. Values representing the global maximum for each dataset are bolded and underlined. Tha A-ROM framework achieved the highest average accuracy of 83.7\% and matched the highest benchmark average AUC of 0.940.}
    \label{fig:benchmark_comp}
\end{figure}

Figure \ref{fig:benchmark_comp} illustrates the performance of A-ROM relative to the MedMNIST v2 benchmarks \cite{yang2023medmnist}, contrasting the results of our optimized  configurations. The optimized model achieved a superior average accuracy of 83.7\% across all datasets, complemented by a highly competitive average AUC of 0.940 that matches the leading benchmark.

\subsection{Few-Shot}

\begin{figure}[ht]
    \centering
    \includegraphics[width=1.0\textwidth]{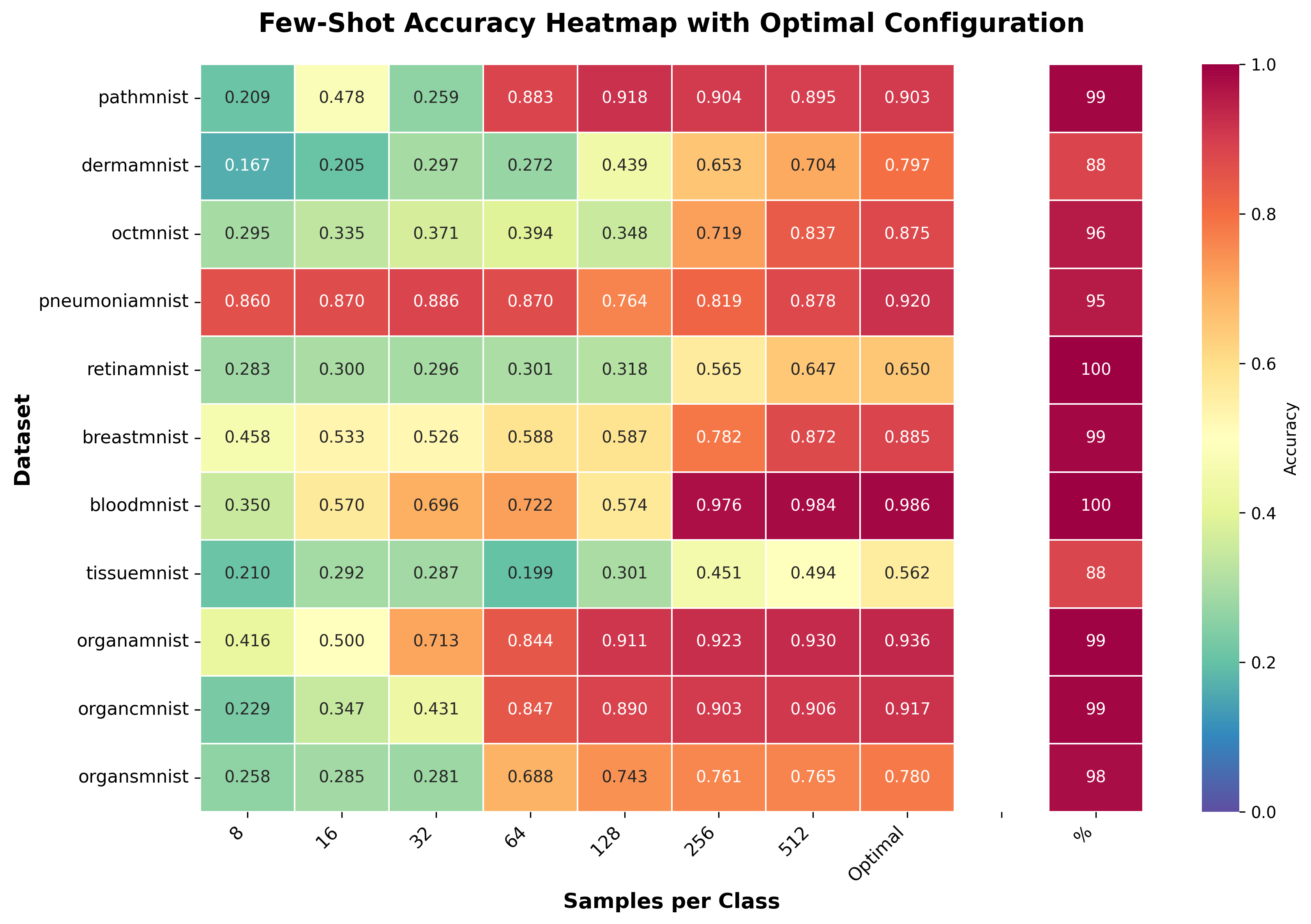}
    \caption{ Classification performance across 11 MedMNIST v2 datasets as a function of training samples per class ($n = 8$ to $512$). The rightmost column denotes the retention rate, representing the percentage of peak accuracy maintained at the 512-sample threshold.}
    \label{fig:few_shot}
\end{figure}

The relationship between training sample availability and classification accuracy is illustrated in Figure \ref{fig:few_shot}. Across the majority of the 11 datasets, a significant performance inflection point occurs at approximately 256 samples per class, suggesting a minimum threshold for establishing a stable, supervised 'concept dictionary.'

As shown in the rightmost column of Figure \ref{fig:few_shot}, the 512-sample configuration retains a high percentage of the accuracy achieved by the fully-sampled optimal model; notably, only two datasets fell below 90\% of their peak performance at this level. These results indicate that as few as 512 labeled samples per class are sufficient for near-optimal performance, highlighting A-ROM’s utility in data-constrained clinical environments where expert labeling is often the primary bottleneck.

\subsection{Interpretability}
\begin{figure}[ht]
    \centering
    \includegraphics[width=1.0\textwidth]{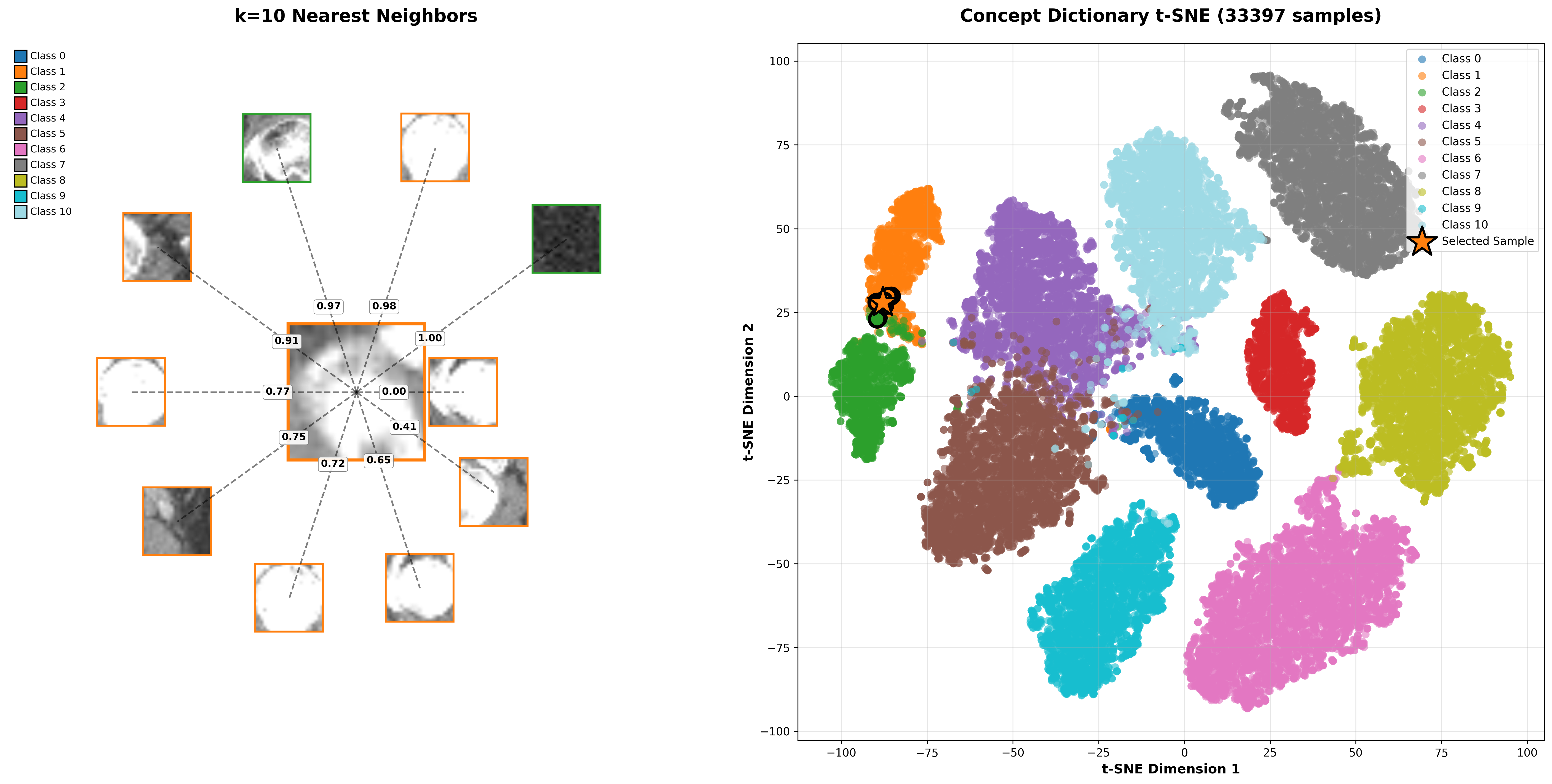}
    \caption{Case-Based Evidence and Manifold Visualization. (Left) A spiral plot illustrating the 10 nearest neighbors to the query sample, ranked by proximity for the organAMNIST dataset. (Right) A t-SNE projection of the latent manifold, situating the test sample and its $k$-nearest neighbors within the global context of the supervised concept dictionary for the organAMNIST dataset.}
    \label{fig:nn_exmple}
\end{figure}

The interpretability of the A-ROM framework is demonstrated in Figure \ref{fig:nn_exmple}. The left panel presents a spiral nearest-neighbor plot, visualizing the training exemplars closest to the query sample alongside their normalized Mahalanobis distances. The right panel contextualizes these local relationships via a global t-SNE projection of the supervised concept dictionary, situating the test sample relative to established class clusters. Together, these visualizations construct a transparent evidentiary chain, enabling clinicians to verify the structural basis of a classification and critically audit cases with low diagnostic confidence.

\section{Discussion}
\label{sec:discussion}
 The results across 11 datasets, 25 network layers, and varying levels of dimensionality reduction invite a deeper analysis of how data characteristics influence the Platonic Representation Hypothesis. A primary area of interest is the relationship between visual features and the framework’s efficacy across different medical modalities.

A-ROM demonstrated exceptional performance relative to benchmarks on DermaMNIST, OCTMNIST, RetinaMNIST, BreastMNIST, BloodMNIST, and OrganCMNIST. These datasets span both color and grayscale modalities but share common visual traits: distinct object boundaries and rich textural information. The high performance on RetinaMNIST is particularly notable given its small sample size, further validating the framework’s label efficiency. These results indicate that the PRH is most effective when imagery possesses clear morphological primitives, enabling the foundation model to resolve the structural regularities required for manifold convergence.

Conversely, the performance gap observed in TissueMNIST, where A-ROM trailed the top benchmark by 14.1\%, is particularly notable given that for all other datasets where A-ROM did not secure the top rank, the margin of difference was within 3.3\%. This outlier suggests a boundary condition for the PRH tied to image acquisition, preprocessing, and training scale. The diffuse confocal fluorescence, compounded by artifacts from $32 \times 32$ upsampling, acts as a low-pass filter suppressing the structural regularities required for manifold convergence. Furthermore, TissueMNIST's volume (165k+ samples) likely enables backpropagation-based models to learn task-specific features. Consequently, while frozen foundation models excel in label-scarce environments with clear morphological primitives, massive datasets may still favor gradient-based optimization.

The framework’s resilience to the extreme dimensionality reduction of LDA further underscores its efficiency. By projecting the combined Alphabet and Vocabulary vectors into a $C-1$ subspace, LDA isolates core class-separability while stripping non-discriminative variance. This multi-stage compression significantly improves the computational feasibility of the final projection, bypassing the overhead of high-dimensional latent spaces. That high predictive accuracy survives an aggressive reduction from 1024 dimensions to fewer than ten underscores the density of the 'Platonic' signal and proves that these features are inherently organized into highly separable spaces.

Overall, these findings establish A-ROM as a robust methodology for the rapid prototyping of diverse medical imaging tasks. By bypassing traditional backpropagation, the framework achieves highly competitive performance with significantly lower computational overhead. Most importantly, by anchoring diagnostic decisions in a scalable, low-dimensional "encoding language," A-ROM provides a human-interpretable evidentiary chain that bridges the gap between deep learning performance and clinical transparency.

\section{Conclusion}
\label{sec:conclusion}
This paper presented the A-ROM framework, which leverages the Platonic Representation Hypothesis to minimize training requirements while producing human-interpretable diagnostic decisions. The framework was rigorously evaluated across all 25 layers of a DINOv2-L/14 backbone, utilizing a wide range of dimensionality reduction settings, and various few-shot learning scenarios.

By bridging Platonic distillation with Aristotelian synthesis, A-ROM achieves a level of conceptual clarity that standard black-box models lack. The results demonstrate that A-ROM is highly competitive across the MedMNIST v2 benchmarks, achieving superior average performance. By bypassing traditional backpropagation based fine-tuning, the framework offers significant practical advantages: an orders-of-magnitude reduction in feature dimensionality, minimal labeled data requirements, and a transparent evidentiary chain. These findings indicate that structured latent 'languages' derived from foundation models provide a robust path toward high-performance, low-overhead, and trustworthy AI for clinical decision support, effectively bridging the gap between state-of-the-art accuracy and bedside interpretability.

\clearpage
\subsubsection{Acknowledgements}
The authors acknowledge the use of Gemini 2.0 Flash (Google) for assistance in refining the narrative structure and editing the manuscript for clarity. Additionally, Claude 3.5 Sonnet (Anthropic) was utilized for accelerating code development, debugging, and data visualizations. All AI-generated content and code were thoroughly reviewed, verified, and tested by the authors to ensure accuracy and originality.

\bibliographystyle{splncs04}
\bibliography{references_v3}

\end{document}